\documentclass[runningheads]{llncs}

\usepackage[T1]{fontenc}
\def\doi#1{\href{https://doi.org/\detokenize{#1}}{\url{https://doi.org/\detokenize{#1}}}}
\usepackage{graphicx}
\usepackage{listings}
\usepackage{amsmath}
\usepackage{amssymb}
\usepackage{bm}
\usepackage{bbm}
\usepackage{caption}
\usepackage{makecell}
\usepackage{multirow}
\usepackage{multicol}
\usepackage{xspace}
\usepackage{url}
\usepackage{wrapfig}
\usepackage[ruled]{algorithm}  
\usepackage[noend]{algpseudocode}
\usepackage{pifont}

\makeatletter
\DeclareRobustCommand\onedot{\futurelet\@let@token\@onedot}
\def\@onedot{\ifx\@let@token.\else.\null\fi\xspace}

\def\eg{\emph{e.g\onedot}} 
\def\ie{\emph{i.e\onedot}} 
\def\cf{\emph{c.f}\onedot}

\makeatother

\begin{document}
\title{Learning Underrepresented Classes from Decentralized Partially Labeled Medical Images}
\titlerunning{Federated Partially Supervised Learning}
%
\author{Nanqing Dong\inst{1}\orcidID{0000-0001-5014-1993} \and Michael Kampffmeyer\inst{2}\orcidID{0000-0002-7699-0405} \and Irina Voiculescu\inst{1}\orcidID{0000-0002-9104-8012}}
%
\authorrunning{N. Dong et al.}
%
\institute{Department of Computer Science, University of Oxford, Oxford, UK \and
UiT The Arctic University of Norway, Tromsø, Norway\\
\email{nanqing.dong@cs.ox.ac.uk}}
\maketitle              
\begin{abstract}
Using decentralized data for federated training is one promising emerging research direction for alleviating data scarcity in the medical domain. However, in contrast to large-scale fully labeled data commonly seen in general object recognition tasks, the local medical datasets are more likely to only have images annotated for a subset of classes of interest due to high annotation costs. In this paper, we consider a practical yet under-explored problem, where underrepresented classes only have few labeled instances available and only exist in a few clients of the federated system. We show that standard federated learning approaches fail to learn robust multi-label classifiers with extreme class imbalance and address it by proposing a novel federated learning framework, FedFew. FedFew consists of three stages, where the first stage leverages federated self-supervised learning to learn \emph{class-agnostic} representations. In the second stage, the decentralized partially labeled data are exploited to learn an energy-based multi-label classifier for the common classes. Finally, the underrepresented classes are detected based on the energy and a \emph{prototype}-based nearest-neighbor model is proposed for few-shot matching.
We evaluate FedFew on multi-label thoracic disease classification tasks and demonstrate that it outperforms the federated baselines by a large margin.

\end{abstract}
\section{Introduction}
\label{sec:intro}
Learning from partially labeled data, or partially supervised learning (PSL), has become an emerging research direction in label-efficient learning on medical images~\cite{zhou2019prior,fang2020multi,shi2021marginal,dong2022towards,dong2022revisiting}. Due to high data collection and annotation costs, PSL utilizes multiple available partially labeled datasets when fully labeled data are difficult to acquire. Here, a partially labeled dataset refers to a dataset with only a specific \emph{true subset} of classes of interest annotated. For example, considering a multi-label thoracic disease task on chest X-ray (CXR) images (\ie~a CXR could contain several diseases at the same time), a pneumonia dataset may only have labels for pneumonia and not the labels for the other diseases of interest.

\begin{figure}
    \centering
    \includegraphics[width=0.8\textwidth]{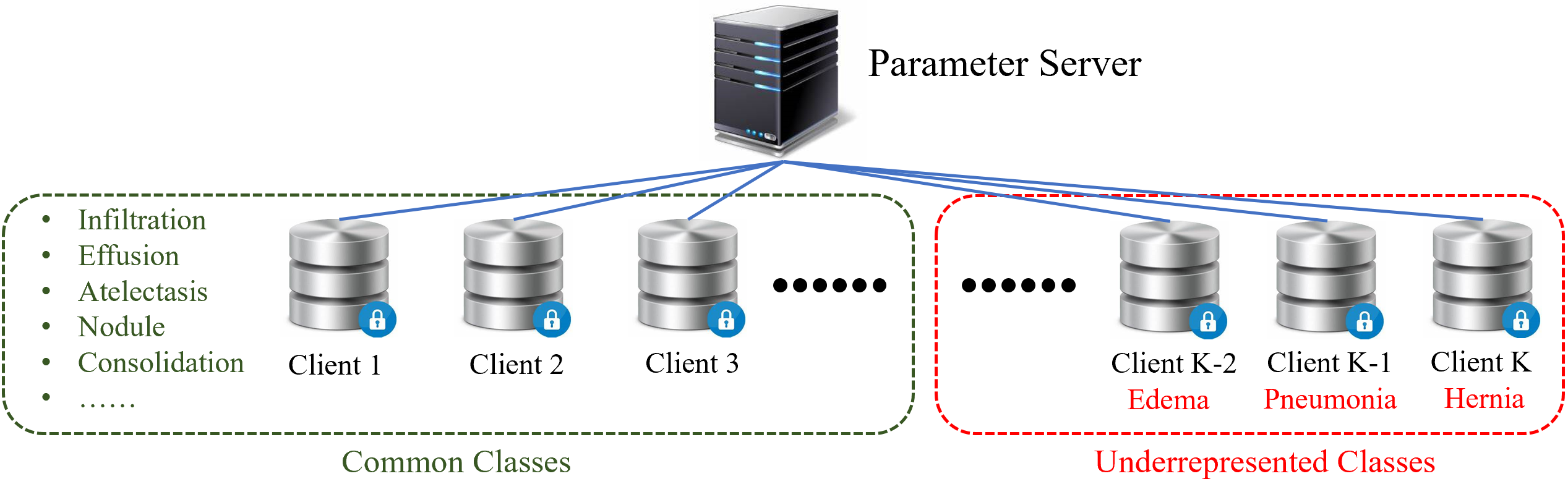}
    \caption{An example problem setup with multi-label thoracic disease classification. In this example, each client in the red dashed box has only one underrepresented class, while the clients in the green dashed box share a set of common classes.
    }
    \label{fig:framework}
\end{figure}

In this work, we extend the discussion of PSL to an unexplored federated setup, where the partially labeled datasets are stored separately in different clients (\eg~hospitals and research institutes). In this work, we denote the common classes (CCs) as the classes with enough labels to learn a multi-label classifier. Meanwhile, in a \emph{open-world} scenario, there are also newly-found or under-examined classes, which tend to have much fewer labeled instances than the CCs. This extreme class imbalance makes learning these underrepresented classes (UCs) a difficult task.
Furthermore, the main assumption of this work is that the CCs and UCs are annotated at disjoint clients, which makes this practical problem even more challenging. While a formal problem setup will be described in Sec.~\ref{sec:prob}, an intuitive illustration of the problem is presented in Fig.~\ref{fig:framework}. Due to decentralization and extreme class imbalance, improving the performance of UCs can inevitably decrease the performance of CCs, while maintaining the performance of CCs might lead to UCs being completely ignored. To the best of our knowledge, this is the first study on decentralized partially labeled rare classes.

\noindent\textbf{Contributions} We formalize this under-explored problem and present FedFew, the first solution to it. FedFew is a three-stage federated learning (FL) framework. Firstly, we utilize federated self-supervised learning (FSSL) to learn \emph{class-agnostic} transferable representations in a pre-training stage.
Secondly, in the fine-tuning stage, we propose an energy-based multi-label classifier that aims to utilize the knowledge from the CCs to learn representations for the UCs. Finally, we utilize a \emph{prototype}-based nearest-neighbor classifier for few-shot matching given only few partially labeled examples for the UCs. We evaluate FedFew on a set of multi-label thoracic disease classification tasks on the Chest-Xray14 dataset~\cite{wang2017chestx} in a simulated federated environment. The empirical results show that the proposed framework outperforms existing methods by a large margin.

\section{Problem Setup}
\label{sec:prob}
\textbf{Task and Data Setup}
The task of interest is multi-label classification (MLC) with $C$ non-mutually exclusive classes of interest, where the decentralized data are stored in a federated system with $K>1$ clients. Let $\mathcal{D}_k$ denote the data stored in client $k \le K$, we have $\mathcal{D}_k \cap \mathcal{D}_l = \emptyset$ for $k \neq l$ and $\{\mathcal{D}_k\}_{k=1}^K$ are non-IID data. For convenience, we define $\mathcal{D}_k = \mathcal{P}_k \cup \mathcal{U}_k$, where $\mathcal{P}_k$ is a \emph{partially labeled} dataset and $\mathcal{U}_k$ is an \emph{unlabeled} dataset. We define $n_k^p = |\mathcal{P}_k|$ and $n_k^u = |\mathcal{U}_k|$.\footnote{$|\cdot|$ is the cardinality of a set.}
We assume that the classes of interest, denoted as $\mathcal{C}$, can be split into two mutually exclusive subsets, namely a set of UCs (denoted as $\mathcal{C}_u \subset \mathcal{C}$), which is also the primary target of this work, and a set of CCs (denoted as $\mathcal{C}_c = \mathcal{C} \setminus \mathcal{C}_u$). For simplicity, we consider a representative case that there are $|\mathcal{C}_u| < K$ clients and each of these clients is annotated for only one UC\footnote{This is the most fundamental case. As a trivial extension, each client could have labels for more than one class and multiple clients could have labels for the same set of classes. The proposed method could be easily adapted to these extensions.}. For the remaining $K - |\mathcal{C}_u|$ clients, we assume that each client has \emph{partially labeled}\footnote{Here, the images with CCs are partially labeled with respect to the missing labels of the UCs, \ie~they are fully labeled if we only consider CCs. The assumption here is that the CCs are diseases with high prevalence, which can be easily collected and diagnosed; but the UCs are rare and only spotted in certain clients.} data for all CCs ($\mathcal{C}_c$). We additionally require $n_i^p \ll n_j^p ~\forall~ i \in \mathcal{C}_u, j~\in \mathcal{C}_c$ to enforce the assumption of UCs. The learning outcome is to leverage the decentralized training data to train an MLC model for $\mathcal{C}$.

\noindent\textbf{Federated Environment Setup}
In addition to $K$ clients, there is a parameter server (PS) \cite{li2014communication} for model aggregation. Let $f_{\theta}$ be the model of interest. In the PS, the parameter set $\theta_0$ is randomly initialized and sent out to $K$ clients as $K$ copies $\{\theta_k\}_{k=1}^K$ for full synchronization. During the federated optimization phase, the client $k$ updates $\theta_k$ by training on $\mathcal{D}_k$ independently for a number of local epochs. Then, the PS aggregates $\{\theta_k\}_{k=1}^K$ collected from $K$ clients to update $\theta_0$. Under the data regulations in the medical domain~\cite{hipaa,gdpr}, we assume that the patients' data (either raw data or encoded data) in a client can not be uploaded to the PS or other clients, \ie~only parameters $\{\theta_k\}_{k=0}^K$ and \emph{metadata} (\eg~the statistics of data) can be exchanged between the PS and the clients.

\section{Method}
\label{sec:method}
In this section, we first provide the preliminaries that FedFew builds on in Sec.~\ref{sec:method:prelim}. The first training stage of FSSL is briefly described in Sec.~\ref{sec:method:fssl}, while the second training stage of energy-based FL with partial labels is described in Sec.~\ref{sec:method:fpsl}. Finally, in Sec.~\ref{sec:method:infer}, we present the prototype-based nearest-neighbor classifier for few-shot matching.

\subsection{Preliminaries}
\label{sec:method:prelim}
\noindent\textbf{FedAvg}
As a seminal FL model, FedAvg~\cite{mcmahan2017communication} aggregates the model weights $\{\theta_k\}_{k=1}^K$ as a weighted average. Mathematically, we have
\begin{equation}
    \theta_0 = \sum_{k=1}^K a_k \theta_k,
    \label{eq:FedAvg}
\end{equation}
where $a_k = \frac{n_k}{n_{tot}}$. The metadata $n_k$ is the number of labeled training examples stored in client $k$ and $n_{tot} = \sum_{k=1}^K n_k$ is the total number of training examples. 

\noindent\textbf{Energy Function}
Given a discriminative neural network classifier $f$, the energy function $E(x; f): \mathbb{R}^{H{\times}W} \rightarrow \mathbb{R}$ maps an image with shape $H{\times}W$ to a scalar, which is also known as \emph{Helmholtz free energy}~\cite{lecun2006tutorial}. The energy is defined as
\begin{equation}
    E(x; f) = -\tau \log \int_{y} \exp^{\frac{f^y(x)}{\tau}},
    \label{eq:energy}
\end{equation}
where $f^y(x)$ is the logit of the $y^{\mathrm{th}}$ class label and $\tau$ is the temperature parameter. 

\subsection{Federated Self-Supervised Learning}
\label{sec:method:fssl}
The first training stage consists of FSSL, where a feature extractor $f_\theta$ is pre-trained to learn class-agnostic representations. Theoretically, multiple existing self-supervised learning frameworks~(\eg~\cite{he2020momentum,chen2020simple,grill2020bootstrap}) could serve as the local backbone. In this work, however, due to its lightweight nature, we leverage SimSiam~\cite{chen2021exploring}. Let $\theta_{0}^{t}$ denote the aggregated model weights in the PS at the end of the $t^{\mathrm{th}}$ training round. Thus, at the beginning of the $t+1^{\mathrm{th}}$ round, the model weights at client $k$ should be synchronized to $\theta_{0}^{t}$. After the local updates of the $t+1^{\mathrm{th}}$ round, the local model weights of client $k$ are now $\theta_{k}^{t+1}$ and, $\theta_{0}^{t+1}$ is computed by applying Eq.~\ref{eq:FedAvg} on $\{\theta_k^{t+1}\}_{k=1}^K$.

\subsection{Energy-Based Federated Learning with Partial Labels}
\label{sec:method:fpsl}
For standard MLC, it is common to use a $C$-dimensional binary vector to encode the label information for a given input. When all binary entries are 0s, the input does not contain any class of interest. However, with limited partial labels of UCs, it is difficult to train an MLC model for $\mathcal{C}$ or $\mathcal{C}_u$ directly. Instead, we first train an MLC model for CCs $\mathcal{C}_c$. In contrast to previous studies~\cite{wang2017chestx,rajpurkar2017chexnet}, we encode the label into a $(C_c+1)$-dimensional vector, where $C_c = |\mathcal{C}_c|$. That is to say, we use an additional dimension (denoted as $0^{\mathrm{th}}$ class\footnote{When the additional dimension is 0, the rest of $C_c$ dimensions should have at least one 1; when the additional dimension is 1, $C_c$ dimensions should all be 0s.}) to specifically determine whether the patient contains any CCs. Note, this $0^{\mathrm{th}}$ class only reflects the information on CCs, as we have no label information for the UCs.

Without loss of generality, let us consider a client $k \in \mathcal{K}_c$ with only $ \mathcal{C}_c$ labeled, where $\mathcal{K}_c$ denotes the clients with the CCs. Given an example $x$ in client $k$ with corresponding partial label $y$, the binary cross-entropy loss is 
\begin{equation}
    \mathcal{L}_{BCE}(x, y) = - \sum_{j \in \{0\} \cup \mathcal{C}_c} y_j \log(f_{\theta}^{j}(x)) + (1 - y_j) \log(1 - f_{\theta}^{j}(x)),
    \label{eq:bce}
\end{equation}
where $f_{\theta}^{j}(x)$ is the probability score for the $j^{\mathrm{th}}$ class. As MLC can be decomposed into multiple binary classification tasks, the energy of $x$ for class $j$ degenerates to $E(x; f_{\theta}^{j}) = -\tau \log(1 + \exp^{\frac{f_{\theta}^{j}(x)}{\tau}})$ (\cf Eq.~\eqref{eq:energy}) and the \emph{joint energy}~\cite{wang2021can} of $x$ is then the sum of energies over all CCs $\mathcal{C}_c$:
\begin{equation}
    E(x, f_\theta) = \sum_{j \in \{0\} \cup \mathcal{C}_c} E(x; f_{\theta}^{j}) = -\sum_{j \in \{0\} \cup \mathcal{C}_c} \tau \log(1 + \exp^{\frac{f_{\theta}^{j}(x)}{\tau}}).
    \label{eq:joint}
\end{equation}
We include a regularization term~\cite{liu2020energy} to penalize the energy of $x$ with a squared hinge loss:
\begin{equation}
    \mathcal{L}_{E_c}(x) = \lambda \|\max(0, E(x; f_\theta) - m_c)\|^2_2,
    \label{eq:E_c}
\end{equation}
where the margin $m_c$ is a hyperparameter chosen empirically to decrease the energy of $x$ and $\lambda$ is a weight hyperparameter. The final optimization goal for client $k \in \mathcal{C}_c$ is to minimize the sum of the two losses:
\begin{equation}
    \mathcal{L}_{c} = \mathcal{L}_{BCE}(x, y) + \mathcal{L}_{E_c}(x).
    \label{eq:total}
\end{equation}
For clients with only UC (\ie~no CCs), we only minimize a regularization term:
\begin{equation}
    \mathcal{L}_{u} = \lambda \|\max(0, m_u - E(x; f_\theta))\|^2_2,
    \label{eq:E_r}
\end{equation}
where the margin $m_u$ is chosen empirically to increase the energy of $x$. Note, Eq.~\eqref{eq:E_c} and Eq.~\eqref{eq:E_r} are both designed to enlarge the \emph{energy gap} between $\mathcal{C}_c$ and $\mathcal{C}_u$. We aggregate the models weights $\{\theta_k\}_{k=1}^K$ via Eq.~\eqref{eq:FedAvg}, where $a_k = \frac{n_k^p}{\sum_{j=1}^K n_j^p}$. The complete pseudo-code is given in Algorithm~\ref{algo:1}.

\begin{algorithm}[t]
    \begin{algorithmic}[1]
    \Statex \textbf{Input:} $\theta_0^0$, $\{\mathcal{P}_k\}_{k=1}^K$, $T_w$, $T$
    \Statex \textbf{Output:} $\theta_0^T$
    \For{$t = 1, 2, \cdots, T_w$} \Comment{Warm up}
    \For{$k \in \mathcal{K}_c$}
    \State{$\theta_k^{t} \leftarrow \theta_0^{t-1}$} \Comment{Synchronize with PS}
    \State{$\theta_k^{t} \leftarrow$ \textit{local\_update}$(\theta_k^{t})$} \Comment{Eq.~\eqref{eq:total}}
    \EndFor
    \State{$\theta_0^{t} \leftarrow \sum_{k \in \mathcal{K}_c} a_{k}^t \theta_{k}^{t}$}\Comment{Aggregate with Eq.~\eqref{eq:FedAvg}}
    \EndFor
    \For{$t = T_w+1, T_w+2, \cdots, T$}
    \For{$k = 1, 2, \cdots, K$}
    \State{$\theta_k^{t} \leftarrow \theta_0^{t-1}$} \Comment{Synchronize with PS}
    \State{$\theta_k^{t} \leftarrow$ \textit{local\_update}$(\theta_k^{t})$} \Comment{Eq.~\eqref{eq:total} or Eq.~\eqref{eq:E_r}}
    \EndFor
    \State{$\theta_0^{t} \leftarrow \sum_{k=1}^K a_{k}^t \theta_{k}^{t}$}\Comment{Aggregate with Eq.~\eqref{eq:FedAvg}}
    \EndFor
    \end{algorithmic}
    \caption{Energy-Based Federated Partially Supervised Training. $T$ is the total number of rounds. $\theta_k^t$ denotes the weights stored in client $k$ at the $t^{\mathrm{th}}$ round.}
    \label{algo:1}
\end{algorithm}

\subsection{Prototype-Based Inference}
\label{sec:method:infer}
After the federated training in Sec.~\ref{sec:method:fpsl}, $f_\theta$ can be directly used as an MLC model to predict CCs.\footnote{Given the $(C_c+1)$-dimensional output vector, we drop the $0^{\mathrm{th}}$ dimension and only use the $C_c$-dimensional vector as the final prediction.} Now, we use the energy (Eq.~\eqref{eq:joint}) to detect UCs, \ie~if the energy of an example is lower than a threshold\footnote{The threshold is chosen to maximize the number of correctly classified training examples.}, then the example is deemed to contain no UCs; if the energy is higher than the threshold, we further match the test example to the nearest neighbor, given few partially labeled examples. However, due to the constraint of data regulations, the training data and the test data are stored in separated clients. Thus, similar to \cite{dong2021federated}, we transfer the metadata of UCs to the PS. Here, the metadata is the mean of the extracted features. For class $c \in \mathcal{C}_u$, we have 
$\bm{\mu}_c^{pos} = \frac{\sum_{i=1}^{n_c^{pos}} g_\theta(x_i^{pos})}{n_c^{pos}},~\bm{\mu}_c^{neg} = \frac{\sum_{i=1}^{n_c^{neg}} g_\theta(x_i^{neg})}{n_c^{neg}},$
where we use $pos$ and $neg$ to denote the positive and negative examples of class $c$, respectively, and use $g_\theta$ to denote the feature extractor. Note, $\bm{\mu}_c^{pos}$ and $\bm{\mu}_c^{neg}$ factually define the \emph{prototypes} in the few-shot learning literature~\cite{snell2017prototypical}. With the \emph{dual}-prototypes for class $c$, we match the test example to the closer one by computing the distance between the features of test example and the two prototypes\footnote{Again, this is a simple case. When there are multiple prototypes collected from different clients for class $c$, majority voting is adopted.}.

\section{Experiments}
\label{sec:exp}
\subsection{Experimental Setup}
\label{sec:exp:setup}
To provide empirical insights into the problem and ensure fair comparisons with the baselines, we share the same experimental setup among all experiments.

\noindent\textbf{Implementation} We explore two network backbones ($f_\theta$), ResNet34 (RN)~\cite{he2016deep} and DenseNet121 (DN)~\cite{huang2017densely}, which are lightweight models commonly used for FL. We use SimSiam~\cite{grill2020bootstrap}, a state-of-the-art SSL framework, to pre-train $f_\theta$ locally, and use a standard Adam~\cite{kingma2015adam} optimizer with fixed learning rate $10^{-3}$ and batch-size 64 for both pre-training and fine-tuning. We set $\tau = 1$ when computing the energy, following~\cite{liu2020energy}. The values of $m_c$ and $m_u$ can be chosen around the mean of energy scores from a model trained without energy-based loss (\eg~warm-up stage in this work) for images from the clients of CCs and the clients of UCs, respectively. In this work, we simply set $m_c = -5$ and $m_u = -25$ to intentionally enlarge the energy gap and we use $\lambda = 0.01$. 
We follow the same data pre-processing and augmentation procedure as~\cite{dong2021federated}. The synchronization and aggregation for federated methods are performed every $10$ epochs. For the second stage, $T_w = 20$ and $T = 100$. All models are implemented in PyTorch (1.10.1) on an NVIDIA Tesla V100.

\noindent\textbf{Data} We use the multi-label dataset ChestX-ray14\footnote{\url{https://nihcc.app.box.com/v/ChestXray-NIHCC}}~\cite{wang2017chestx} and adopt its default batch splits to ensure reproducibility. Based on the label statistics of the dataset, we choose \emph{edema}, \emph{pneumonia}, and \emph{hernia} as the three UCs and use the remaining 11 classes as CCs. Note, most CXR images do not contain any diseases. We use 6 batches\footnote{We use batch 2 to 7. Each batch has $10^4$ CXR images and similar label distributions.} to simulate the $K = 6$ clients, where we use the first three batches to simulate the partially labeled datasets for CCs, where we randomly sample $n_k^p = 5{\times}10^3$ images to keep partial labels for CCs. Each of the remaining three batches contains one of the three UCs. We sample $10$ positive examples and $90$ negative examples to simulate the class imbalance for UCs, \ie~$n_k^p = 100$. See Fig.~\ref{fig:framework} for an illustration of the class assignment.
From the remaining batches, we hold out 100 positive and 100 negative examples for each class as the test set.

\subsection{Results}
\label{sec:exp:result}
\noindent\textbf{Empirical Analysis of FSSL} Following previous SSL studies, we examine the FSSL performance via the \emph{linear classification protocol}~\cite{he2020momentum,chen2020simple,grill2020bootstrap}. Similar to~\cite{dong2021federated}, we fix all the weights of $f_\theta$ except the last layer and only fine-tune the last layer on a public pneumonia dataset\footnote{\url{https://data.mendeley.com/datasets/rscbjbr9sj/2}}~\cite{kermany2018identifying}. In this dataset, there are three mutually exclusive classes, \emph{normal}, \emph{bacteria pneumonia}, and \emph{virus pneumonia}. We randomly split the images of each class into two halves as the training and test sets. We use the test accuracy as the \emph{proxy} measure to assess the representation learning performance. Firstly, we provide a counter-intuitive observation that more disease images might not improve the performance. We create two clients with Chest-Xray14 data, where one client contains $10^4$ images without any diseases and the other client contains $r{\times}10^4$ images with various diseases. The results in Table~\ref{tab:pretrain} show that FedAvg does not always benefit from large $r$ and it might be unnecessary to collect a large number of images with related diseases for pre-training. Secondly, we examine the impact of the training epochs for federated SSL in Fig.~\ref{fig:epochs}, where we pre-train $f_\theta$ on the 6 clients described in Sec.~\ref{sec:exp:setup}. In contrast to the empirical findings collected from general images~\cite{he2020momentum,chen2020simple,grill2020bootstrap,chen2021exploring}, more epochs will not lead to diminishing performance gain but decreasing results. We will use DenseNet121 as the default network and use the pre-trained weights with 100 epochs in the following experiments. 

\begin{table}[t]
    \captionsetup{type=table}
    \begin{minipage}{0.45\textwidth}
    \centering
    {\scriptsize
    \begin{tabular}{|c|c|c|c|c|c|c|}\hline
    \multirow{2}{*}{Epoch} & \multicolumn{2}{c|}{$r = 1$} & \multicolumn{2}{c|}{$r = 0.5$} & \multicolumn{2}{c|}{$r = 0.1$}\\\cline{2-7}
    & RN & DN & RN & DN & RN & DN \\ \hline
    100 & 61.69 & 70.47 & 69.34 & 71.94 & 65.92 & 73.81\\ \hline
    200 & 62.96 & 71.22 & 68.32 & 71.36 & 65.24 & 74.43 \\ \hline
    \end{tabular}
    \caption{Impact of class imbalance on FSSL. We report the mean accuracy over three random seeds.}
    \label{tab:pretrain}
    }
    \end{minipage}
    \begin{minipage}{0.5\textwidth}
    \centering
    \captionsetup{type=figure}
    \includegraphics[width=0.85\textwidth]{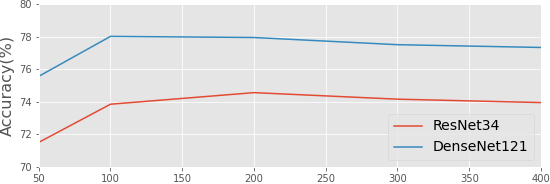}
    \caption{Impact of the number of training epochs on FSSL.}
    \label{fig:epochs}
  \end{minipage}
\end{table}

\noindent\textbf{Evaluation of FedFew} Following the experimental setup in Sec.~\ref{sec:exp:setup}, we evaluate FedFew against a few seminal baselines from two aspects, \ie~we aim to achieve high accuracy on UCs while maintaining robust performance on CCs. The first baseline is a standard MLC model~\cite{rajpurkar2017chexnet}, which is trained with FedAvg on decentralized partially labeled data of all 14 classes and weighted binary cross-entropy~\cite{rajpurkar2017chexnet}. 
We use \emph{MLC w/o FSSL} and \emph{MLC w/ FSSL} to differentiate whether $f_\theta$ is pre-trained with FSSL. The second baseline is a hypothetical local nearest-neighbor classifier (\emph{NN}) that does not use prototypes, where $g_\theta$ is either pre-trained with FSSL alone or further fine-tuned with \emph{MLC w/ FSSL}. Note, \emph{NN} can not be used in a federated system due to data regulations. For FedFew, we use \emph{EBM} to denote the energy-based loss in the training. The results on UCs are presented in Table~\ref{tab:mlc} and include the mean accuracy (A), precision (P), recall (R), and F-1 (F) score over three random seeds. 
Note, standard MLC models fail to detect any images of the UCs due to extreme class imbalance. With only prototypes (as only metadata can be transferred to the PS), NNs struggle to improve over random guessing. FedFew (\emph{w/o EBM}) outperforms the two baselines by a large margin while \emph{EBM} further improves the performance of FedFew with higher precision. Similar to Table~\ref{tab:mlc}, we report the mean \emph{area under the receiver operating characteristic} (AUROC) for the 11 CCs over three runs in Table~\ref{tab:mlc_cc}. We find that FSSL can boost the performance for standard MLC. While FedFew (\emph{w/o EBM}) and FedFew (\emph{w/ EBM}) both outperform \emph{MLC w/ FSSL}, including \emph{EBM} slightly improves the performance on CCs. We conjecture that \emph{EBM} can play a role as regularization in the federated training. By combining the results from both Table~\ref{tab:mlc} and Table~\ref{tab:mlc_cc}, we conclude that FedFew (\emph{w/ EBM}) provides a robust solution to the problem of interest.

\noindent\textbf{Analysis on Energy} As an ablation study, we visualize the energy density plots between images with and without UCs in Fig.~\ref{fig:energy}, which demonstrates that including \emph{EBM} in the training does increase the energy gap, thus leading to improved performance.

\noindent\textbf{Distance Metric} We consider three distance metrics for the nearest-neighbor matching is Sec.~\ref{sec:method:infer}, which are cosine distance~\cite{vinyals2016matching}, Euclidean distance~\cite{snell2017prototypical}, and earth mover's distance~\cite{zhang2020deepemd} (computed with Sinkhorn-Knopp algorithm~\cite{cuturi2013sinkhorn}). We choose the cosine distance based on its empirical robustness. It achieves an average F-1 score over the UCs of 0.67, while the Euclidean distance obtains 0.6 and the earth mover's distance obtains 0.43.

\begin{table}[t]
  \centering
  \setlength{\tabcolsep}{0.2em}
  {\scriptsize
  \begin{tabular}{|l|c|c|c|c|c|c|c|c|c|c|c|c|}
    \hline
    \multirow{2}{*}{Method} & \multicolumn{4}{c|}{Edema} & \multicolumn{4}{c|}{Pneumonia} & \multicolumn{4}{c|}{Hernia} \\\cline{2-13}
     & A & P & R & F & A & P & R & F & A & P & R & F \\\hline
     MLC w/o FSSL & 0.50 & 0.00 & 0.00 & 0.00 & 0.50 & 0.00 & 0.00 & 0.00 & 0.50 & 0.00 & 0.00 & 0.00 \\\hline
     MLC w/ FSSL & 0.50 & 0.00 & 0.00 & 0.00 & 0.50 & 0.00 & 0.00 & 0.00 & 0.50 & 0.00 & 0.00 & 0.00 \\\hline
     NN (FSSL) & 0.53 & 0.53 & 0.50 & 0.52 & 0.51 & 0.51 & 0.46 & 0.48 & 0.50 & 0.50 & 0.54 & 0.52 \\\hline
     NN (MLC w/ FSSL) & 0.53 & 0.53 & 0.50 & 0.52 & 0.51 & 0.51 & 0.46 & 0.48 & 0.50 & 0.50 & 0.54 & 0.52 \\\hline
     FedFew w/o EBM & 0.71 & 0.85 & 0.50 & 0.63 & 0.69 & 0.84 & 0.46 & 0.59 & 0.72 & 0.83 & 0.54 & 0.65 \\\hline
     FedFew w/ EBM & \textbf{0.75} & \textbf{1.00} & \textbf{0.50} & \textbf{0.67} & \textbf{0.73} & \textbf{1.00} & \textbf{0.46} & \textbf{0.63} & \textbf{0.77} & \textbf{1.00} & \textbf{0.54} & \textbf{0.70}\\\hline
  \end{tabular}
  }
  \caption{Performance comparison on the UCs. The standard MLC and NN models fail to predict the UCs. A, P, R, and F denote the mean accuracy, precision, recall, and F1-score over three random seeds, respectively.}
  \label{tab:mlc}
\end{table}

\begin{table}[t]
    \captionsetup{type=table}
    \begin{minipage}{0.45\textwidth}
    \centering
    {\scriptsize
    \begin{tabular}{|l|c|}\hline
    Method & AUROC \\\hline
    MLC w/o FSSL & 0.6905 \\\hline
    MLC w/ FSSL & 0.7227 \\\hline
    FedFew w/o EBM & 0.7423 \\\hline
    FedFew w/ EBM & \textbf{0.7479} \\\hline
    \end{tabular}
    \caption{Performance comparison on the CCs. AUROC denotes the mean AUROC over three random seeds.}
    \label{tab:mlc_cc}
    }
    \end{minipage}
  \begin{minipage}{0.52\textwidth}
    \centering
    \captionsetup{type=figure}
    \includegraphics[width=\textwidth]{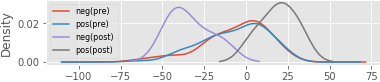}
    \caption{Energy gaps between positive and negative cases before (pre) and after (post) \emph{EBM} training in Client 4.}
    \label{fig:energy}
  \end{minipage}
\end{table}

\section{Conclusion}
In this work, we raise awareness of an under-explored problem, namely the learning of underrepresented classes from decentralized partially labeled medical images. We not only provide a solution to this novel problem but also provide the first empirical understanding of federated partially supervised learning with extreme class imbalance, a new research direction on label-efficient learning.
%
%
%
\bibliographystyle{splncs04}
\bibliography{refs}

\begin{thebibliography}{10}
\providecommand{\url}[1]{\texttt{#1}}
\providecommand{\urlprefix}{URL }
\providecommand{\doi}[1]{https://doi.org/#1}

\bibitem{chen2020simple}
Chen, T., Kornblith, S., Norouzi, M., Hinton, G.: A simple framework for
  contrastive learning of visual representations. In: ICML. pp. 1597--1607.
  PMLR (2020)

\bibitem{chen2021exploring}
Chen, X., He, K.: Exploring simple siamese representation learning. In: CVPR.
  pp. 15750--15758 (2021)

\bibitem{cuturi2013sinkhorn}
Cuturi, M.: Sinkhorn distances: Lightspeed computation of optimal transport.
  In: NIPS. vol.~26, pp. 2292--2300 (2013)

\bibitem{dong2022towards}
Dong, N., Kampffmeyer, M., Liang, X., Xu, M., Voiculescu, I., Xing, E.: Towards
  robust partially supervised multi-structure medical image segmentation on
  small-scale data. Applied Soft Computing p. 108074 (2022)

\bibitem{dong2021federated}
Dong, N., Voiculescu, I.: Federated contrastive learning for decentralized
  unlabeled medical images. In: MICCAI. pp. 378--387. Springer (2021)

\bibitem{dong2022revisiting}
Dong, N., Wang, J., Voiculescu, I.: Revisiting vicinal risk minimization for
  partially supervised multi-label classification under data scarcity. In: CVPR
  Workshops. pp. 4212--4220 (2022)

\bibitem{gdpr}
{European Commission}: General data protection regulation (2016),
  \url{https://ec.europa.eu/info/law/law-topic/data-protection/data-protection-eu_en}

\bibitem{fang2020multi}
Fang, X., Yan, P.: Multi-organ segmentation over partially labeled datasets
  with multi-scale feature abstraction. IEEE TMI  (2020)

\bibitem{grill2020bootstrap}
Grill, J.B., Strub, F., Altch{\'e}, F., Tallec, C., Richemond, P., Buchatskaya,
  E., Doersch, C., Pires, B., Guo, Z., Azar, M., et~al.: Bootstrap your own
  latent: A new approach to self-supervised learning. In: NIPS. vol.~33, pp.
  21271--21284 (2020)

\bibitem{he2020momentum}
He, K., Fan, H., Wu, Y., Xie, S., Girshick, R.: Momentum contrast for
  unsupervised visual representation learning. In: CVPR. pp. 9729--9738 (2020)

\bibitem{he2016deep}
He, K., Zhang, X., Ren, S., Sun, J.: Deep residual learning for image
  recognition. In: CVPR. pp. 770--778 (2016)

\bibitem{huang2017densely}
Huang, G., Liu, Z., Van Der~Maaten, L., Weinberger, K.Q.: Densely connected
  convolutional networks. In: CVPR. pp. 4700--4708 (2017)

\bibitem{kermany2018identifying}
Kermany, D.S., Goldbaum, M., Cai, W., Valentim, C.C., Liang, H., Baxter, S.L.,
  McKeown, A., Yang, G., Wu, X., Yan, F., et~al.: Identifying medical diagnoses
  and treatable diseases by image-based deep learning. Cell  \textbf{172}(5),
  1122--1131 (2018)

\bibitem{kingma2015adam}
Kingma, D.P., Ba, J.: Adam: A method for stochastic optimization. In: ICLR
  (2015)

\bibitem{lecun2006tutorial}
LeCun, Y., Chopra, S., Hadsell, R., Ranzato, M., Huang, F.: A tutorial on
  energy-based learning. Predicting Structured Data  \textbf{1}(0) (2006)

\bibitem{li2014communication}
Li, M., Andersen, D.G., Smola, A.J., Yu, K.: Communication efficient
  distributed machine learning with the parameter server. In: NIPS. pp. 19--27
  (2014)

\bibitem{liu2020energy}
Liu, W., Wang, X., Owens, J., Li, Y.: Energy-based out-of-distribution
  detection. In: NIPS. vol.~33, pp. 21464--21475 (2020)

\bibitem{mcmahan2017communication}
McMahan, B., Moore, E., Ramage, D., Hampson, S., Aguera~y Arcas, B.:
  Communication-efficient learning of deep networks from decentralized data.
  In: AISTATS. pp. 1273--1282. PMLR (2017)

\bibitem{rajpurkar2017chexnet}
Rajpurkar, P., Irvin, J., Zhu, K., Yang, B., Mehta, H., Duan, T., Ding, D.,
  Bagul, A., Langlotz, C., Shpanskaya, K., et~al.: Chexnet: Radiologist-level
  pneumonia detection on chest x-rays with deep learning. arXiv preprint
  arXiv:1711.05225  (2017)

\bibitem{shi2021marginal}
Shi, G., Xiao, L., Chen, Y., Zhou, S.K.: Marginal loss and exclusion loss for
  partially supervised multi-organ segmentation. Medical Image Analysis p.
  101979 (2021)

\bibitem{snell2017prototypical}
Snell, J., Swersky, K., Zemel, R.: Prototypical networks for few-shot learning.
  In: NIPS. pp. 4077--4087 (2017)

\bibitem{hipaa}
{US Department of Health and Human Services}: Health insurance portability and
  accountability act (2017),
  \url{https://www.cdc.gov/phlp/publications/topic/hipaa.html}

\bibitem{vinyals2016matching}
Vinyals, O., Blundell, C., Lillicrap, T., Wierstra, D., et~al.: Matching
  networks for one shot learning. In: NIPS. pp. 3630--3638 (2016)

\bibitem{wang2021can}
Wang, H., Liu, W., Bocchieri, A., Li, Y.: Can multi-label classification
  networks know what they don’t know? In: NIPS. vol.~34 (2021)

\bibitem{wang2017chestx}
Wang, X., Peng, Y., Lu, L., Lu, Z., Bagheri, M., Summers, R.M.: Chestx-ray8:
  Hospital-scale chest x-ray database and benchmarks on weakly-supervised
  classification and localization of common thorax diseases. In: CVPR. pp.
  2097--2106 (2017)

\bibitem{zhang2020deepemd}
Zhang, C., Cai, Y., Lin, G., Shen, C.: Deepemd: Few-shot image classification
  with differentiable earth mover's distance and structured classifiers. In:
  CVPR. pp. 12203--12213 (2020)

\bibitem{zhou2019prior}
Zhou, Y., Li, Z., Bai, S., Wang, C., Chen, X., Han, M., Fishman, E., Yuille,
  A.L.: Prior-aware neural network for partially-supervised multi-organ
  segmentation. In: ICCV. pp. 10672--10681 (2019)

\end{thebibliography}

\end{document}